\documentclass[10pt,twocolumn,letterpaper]{article}

\usepackage{cvpr}              
\usepackage{amssymb}
\newcommand{\newcheckmark}{\dingbat@sym{'104}}
\usepackage{dsfont}
\usepackage{float}
\usepackage{algorithm}  
\usepackage{algpseudocode}
\usepackage{algorithmicx}
\usepackage[T1]{fontenc}

\definecolor{cvprblue}{rgb}{0.21,0.49,0.74}
\usepackage[pagebackref,breaklinks,colorlinks,allcolors=cvprblue]{hyperref}

\title{A Lightweight Moment Retrieval System with Global Re-Ranking and Robust Adaptive Bidirectional Temporal Search}

\author{
\thanks{All authors contributed equally to this paper. \\ 
This research is fully supported by AI VIETNAM \cite{aivietnamVit}.}
Tinh-Anh Nguyen-Nhu$^{1*}$ \quad Huu-Loc Tran$^{2*}$ \quad  Nguyen-Khang Le$^{3*}$ \quad Minh-Nhat Nguyen$^{4*}$ \\Tien-Huy Nguyen$^{2}$ \quad Hoang-Long Nguyen-Huu$^{2}$ \quad Huu-Phong Phan-Nguyen$^{2}$ \\ \quad Huy-Thach Pham $^{5}$\quad Quan Nguyen $^{6}$\quad Hoang M. Le $^{7}$\quad Quang-Vinh Dinh $^{8}$\\ \\
$^{1}$ Ho Chi Minh University of Technology, VNU-HCM, Vietnam \\
$^{2}$ University of Information Technology, VNU-HCM, Vietnam \\
$^{3}$ Ho Chi Minh University of Science, VNU-HCM, Vietnam \\
$^{4}$ University of Economics Ho Chi Minh City, Ho Chi Minh, Vietnam \\
$^{5}$ Georgia State University, USA \\
$^{6}$ Posts and Telecommunications Institute of Technology, Hanoi, Vietnam \\
$^{7}$ York University, Canada\\
$^{8}$ AI VIETNAM Lab \\
{\tt\small 22520567@gm.uit.edu.vn}}

\begin{document}
\maketitle

\begin{abstract}

The exponential growth of digital video content has posed critical challenges in moment-level video retrieval, where existing methodologies struggle to efficiently localize specific segments within an expansive video corpus. Current retrieval systems are constrained by computational inefficiencies, temporal context limitations, and the intrinsic complexity of navigating video content. In this paper,  we address these limitations through a novel Interactive Video Corpus Moment Retrieval framework that integrates a SuperGlobal Reranking mechanism and Adaptive Bidirectional Temporal Search (ABTS), strategically optimizing query similarity, temporal stability, and computational resources. By preprocessing a large corpus of videos using a keyframe extraction model and deduplication technique through image hashing, our approach provides a scalable solution that significantly reduces storage requirements while maintaining high localization precision across diverse video repositories.

\end{abstract}    
\section{Introduction}
\label{sec:intro}

Recent advances in deep learning and computer vision have led to remarkable performance across a wide range of tasks, including visual question answering, object detection, recognition, and domain adaptation \cite{nguyen2024improvinggeneralizationvisualreasoning, 10661057, nguyen2025enhancing, nguyen2024emotic, ngo2024dual, nguyen5109180mv}. Besides,  with the rapid expansion of online video platforms, Video Corpus Moment Retrieval (VCMR) faces significant challenges, particularly in handling long videos with redundant content, leading to apply deep learning method to solve problems. VCMR involves identifying specific moments within videos from a large repository, typically combining Video Retrieval and Single Video Moment Retrieval (SVMR) \cite{zhang2021video, hou2024event, li2020hero, zhang2025video}. However, long videos with irrelevant segments degrade retrieval performance and increase storage resources \cite{yin2024text}. Additionally, text-to-video models struggle to localize moments effectively, as excessive frames obscure key features \cite{wu2023empirical}.

Recent methods utilize keyframe-based retrieval \cite{phan2023doppelsearch, le2023enhancing} to reduce processing costs, yet ignoring temporal structure hinders precise boundary detection. Moreover, retrieval noise persists due to ambiguous queries and overlapping content. Reranking techniques refine results by incorporating temporal consistency, evolving from feature matching \cite{arandjelovic2012three, philbin2007object} to Transformer-based models \cite{tan2021instance}, though computational costs remain a challenge.

To address these limitations, we propose an efficient VCMR framework that combines keyframe-based image retrieval with temporal refinement. Our method significantly reduces processing and storage costs while maintaining high localization accuracy. This work introduces the following key contributions:
\begin{itemize}
    \item \textbf{Rerank Module:} We leverage a novel reranking mechanism called SuperGlobal Reranking \cite{shao2023global}. By refining the initial candidate moments, the approach combines both stages of reranking into a single global stage, effectively reducing both the memory footprint and computational time without sacrificing the overall performance.
    \item \textbf{Temporal Search:} We propose Adaptive Bidirectional  Temporal Search (ATBS), a novel method for enhancing moment retrieval by jointly optimizing query similarity and temporal stability. Unlike traditional similarity-based approaches that often suffer from boundary misalignment, ATBS employs a bidirectional search strategy to accurately identify the start and end frames of a moment. Additionally, we introduce stability weighting, which prioritizes boundaries that exhibit higher temporal coherence and consistency, ensuring a more reliable segmentation.
\end{itemize}






\section{Related work}
\label{sec:related}

\textbf{Video Corpus Moment Retrieval (VCMR)} aims to locate specific video segments that align with textual queries, leveraging a variety of innovative approaches. Research spans supervised~\cite{jung2022modal,gao2021fast,cui2022video}, weakly-supervised~\cite{cui2022video,lin2022weakly}, and zero-shot paradigms leveraging multimodal large language models~\cite{zhang2024momentgpt}. Efficiency challenges are addressed through fast retrieval frameworks~\cite{gao2021fast} and cross-modal common spaces. Annotation cost reduction techniques include "glance annotation"~\cite{cui2022video} and pretraining on unlabeled videos~\cite{anonymous2023vidmorp}. Performance improvements come from modal-specific query generation~\cite{jung2022modal}, transformer architectures~\cite{jones2024lddetr}, cross-modal interaction~\cite{jung2022modal,gao2021fast}, graph neural networks~\cite{wang2023graph}, reinforcement learning~\cite{wu2022reinforcement}, semantic-conditioned modulation~\cite{mun2022local}, boundary-aware prediction~\cite{liu2023boundary}, sentence reconstruction~\cite{chen2023semantic}, tree LSTM structures~\cite{wang2023tree}, and unified timestamp localization frameworks~\cite{zhao2023timeloc}.\\

\indent\textbf{Video Question Answering (VideoQA)} has progressed significantly, beginning with early methods that adapted image-based QA by incorporating temporal modeling, attention across frames, and memory networks to capture dynamic context~\cite{jang2017tgifqa,gao2018motion,jang2019video,xu2017gradually, nguyen2024improvinggeneralizationvisualreasoning}. Multimodal fusion strategies then grew more powerful, leveraging cross-modal co-attention mechanisms, hierarchical video representations, graph convolutional networks, and even transformer-based encoders to jointly model visual and linguistic information for improved alignment and understanding~\cite{li2019beyond,jiang2020divide,park2021bridge}. To handle complex reasoning, later works introduced relational modules like multi-step attention, spatio-temporal scene graphs, and neuro-symbolic frameworks~\cite{song2018multi,yi2020clevrer,mao2022dynamic,xiao2022video}. Recently, the focus has shifted to scalable, generalizable models using large-scale video–language pre-training and synthetic data, enabling unified transformer-based architectures to achieve strong zero-shot or few-shot performance across diverse datasets, domains, and question types with minimal task-specific fine-tuning~\cite{lei2021clipbert,yang2021justask,zellers2021merlot,yang2022zeroshot,wu2021transferring,li2022learning,li2020hero}. \\
\indent\textbf{Interactive Video Retrieval (IVR)} enables human-machine collaboration to iteratively refine video search results, addressing the semantic gap in automated methods. Early systems used relevance feedback based on low-level features~\cite{rui1998relevance,christel2005improving}, later evolving into embedding-based models that adapt to user input across sessions~\cite{arandjelovic2012query,jain2016query}. Language-driven IVR introduced natural language commands and follow-ups for dynamic refinement~\cite{liu2021dynamic,lee2021video,alam2022revisiting}, powered by vision-language models like CLIP and VideoBERT for flexible query understanding~\cite{fang2023clip2video,bain2021frozen}. Reinforcement learning and iterative grounding further improved retrieval by modeling user intent and refining temporal-spatial scopes~\cite{das2017learning,seo2021reinforced,miech2020end,akbari2021vatt}. Enhancements like visual dashboards, explainability, and few-shot personalization improved usability~\cite{papadopoulos2017click,law2018vision,liu2023hyper}, while active learning reduced annotation costs~\cite{agarwal2020active,chen2022few}. Large-scale benchmarks show IVR outperforms automated systems in precision-demanding tasks~\cite{lokoc2019interactive,smeaton2006evaluation,lokoc2023interactive}, pointing toward a future of adaptive, multimodal, feedback-driven retrieval systems. Recent systems from the Video Browser Showdown (VBS) demonstrate the growing capabilities of interactive video search. diveXplore~\cite{Leibetseder2020DiveXplore} supports diverse multimodal queries and collaborative exploration, while VISIONE and vitrivr~\cite{Amato2021VISIONE,Sauter2020Vitrivr}  leverage scalable indexing and content-based retrieval with rich query support. Exquisitor and VIRET~\cite{Jonsson2020Exquisitor,Lokoc2019VIRET} enhance relevance feedback and deep model-based annotation, enabling more precise and flexible search interactions.

\section{Methodology}
\label{sec:method}
\begin{figure*}[t]
    \centering
    \includegraphics[width=0.95\linewidth]{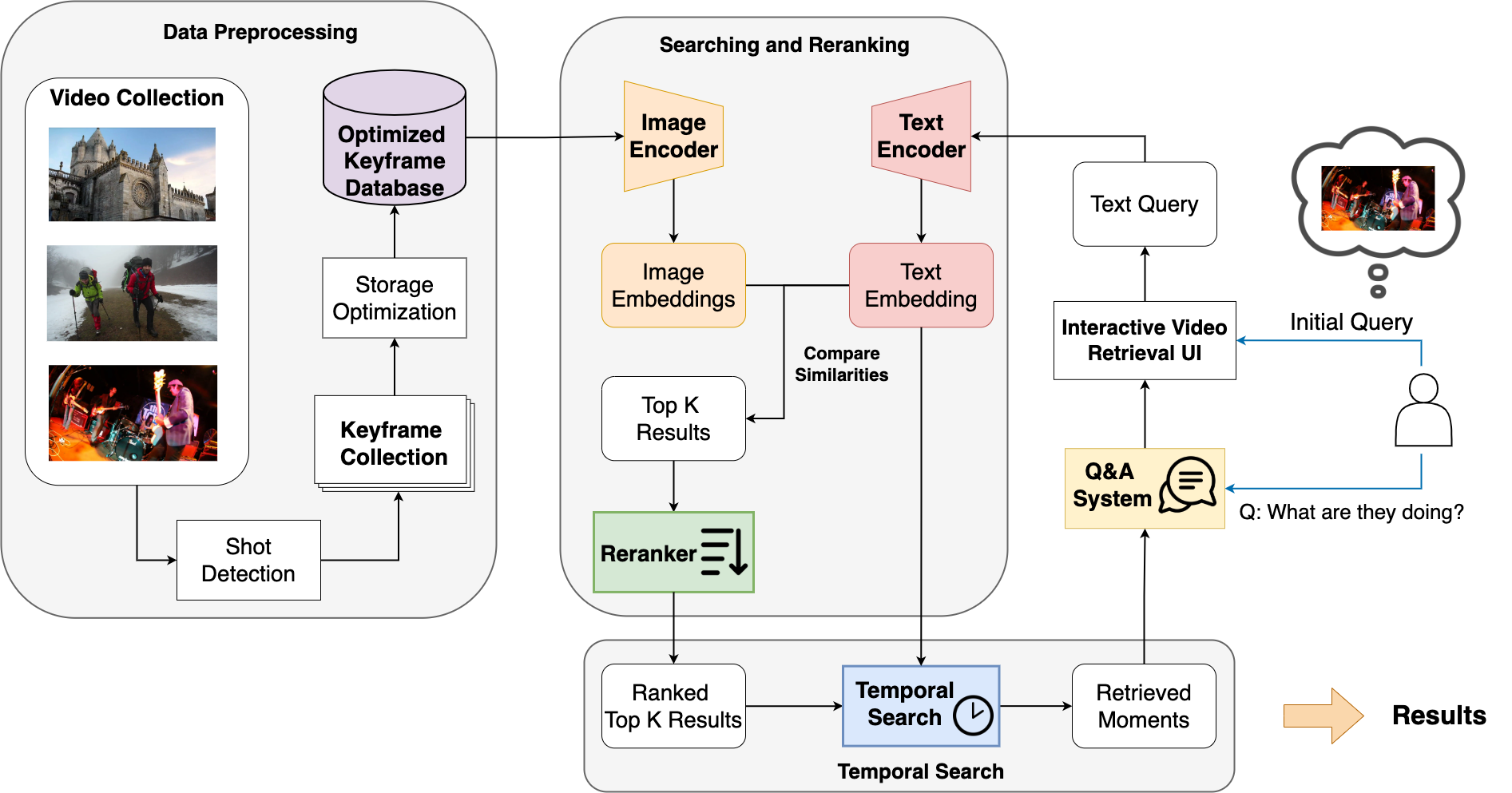}
    \caption{Overview of our \textbf{GRAB — Global Re-ranking and Adaptive Bidirectional} search system. The user begins by entering a natural language query to search for semantically relevant keyframes in a preprocessed video corpus. a) In \textbf{Section} \ref{subsec:data_preprocessing} data preprocessing, raw videos are segmented using shot detection, and representative keyframes are extracted and deduplicated to form a storage-efficient and visually diverse index. b) In \textbf{Section} \ref{subsec:search_rerank} Embedding-based searching and reranking, the user query is embedded and compared against the keyframe database using FAISS for fast retrieval, followed by SuperGlobal Reranking to refine the results. The user then selects a pivot frame from the top-ranked results. c) In \textbf{Section} \ref{subsec:temporal_search}, Adaptive Bidirectional Temporal Search identifies precise start and end boundaries based on semantic similarity and temporal stability. The interface supports interactive refinement and QA-based boundary validation.}
    \vspace{-1.5em}
    \label{fig:system_overview}
\end{figure*}

To overcome the limitations of existing video moment retrieval systems,  especially in handling long, untrimmed videos with high redundancy and weak temporal precision, we propose \textbf{GRAB} - a modular framework that integrates efficient keyframe-based search with adaptive temporal localization. Our system is designed to significantly reduce computational overhead while improving retrieval accuracy and boundary localization. 
As illustrated in \textbf{Figure}~\ref{fig:system_overview}, our proposed framework comprises three core stages. First, in Data Preprocessing (\textbf{Section}~\ref{subsec:data_preprocessing}), we perform shot detection and extract a compact set of keyframes using a perceptual hashing-based deduplication strategy, resulting in a storage-optimized keyframe database. Next, in Searching and Reranking (\textbf{Section}~\ref{subsec:search_rerank}), user queries are embedded and matched against keyframe embeddings using FAISS for efficient similarity search, followed by a reranking stage that refines results through contextual feature enhancement. Finally, Temporal Search (\textbf{Section}~\ref{subsec:temporal_search}) takes the top retrieved frames and performs bidirectional localization using an adaptive scoring mechanism that balances semantic similarity with temporal stability to identify precise start and end timestamps.

\subsection{System Overview}

\begin{figure}
    \centering
    \fbox{\includegraphics[width=1\linewidth]{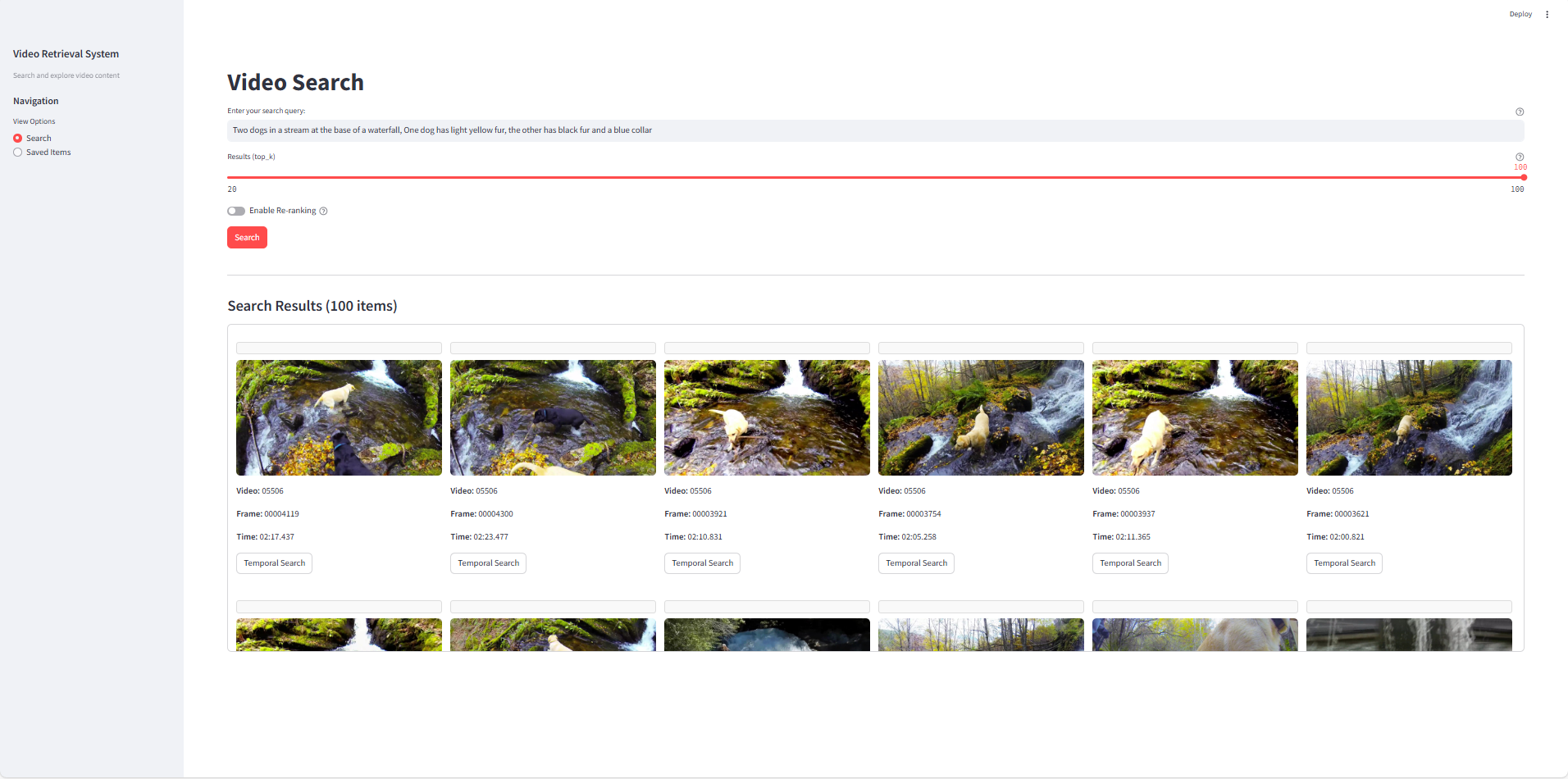}}
    \caption{User Interface of Our Interactive Video Corpus Moment Retrieval System.}
    \label{fig:user_interface}
\end{figure}

\begin{figure}[t]
    \centering
    \begin{subfigure}{0.4\textwidth}
        \centering
        \fbox{\includegraphics[width=\linewidth]{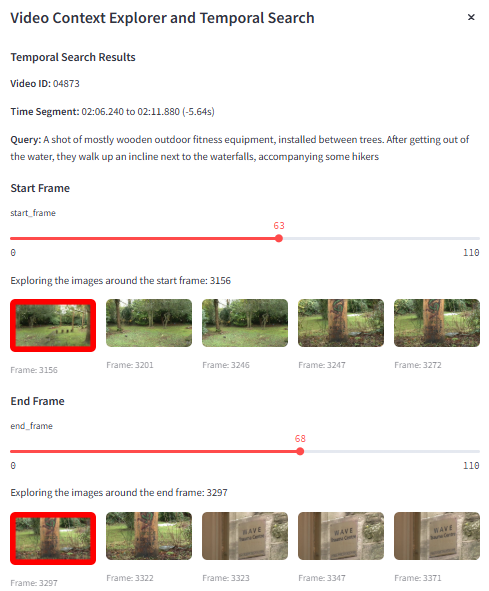}}
        \caption{Moment Selection}
        \label{fig:moment_selection}
    \end{subfigure}
    \hfill
    \begin{subfigure}{0.4\textwidth}
        \centering
        \fbox{\includegraphics[width=\linewidth]{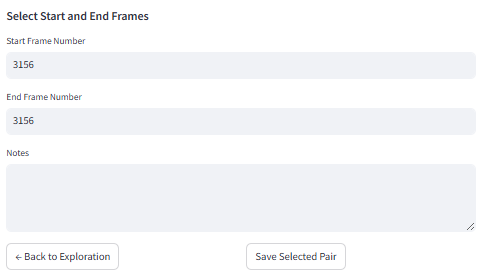}}
        \caption{QA and Boundary Selection}
        \label{fig:qa_boundary}
    \end{subfigure}
    \caption{Visualization of different interaction components: (a) Moment Exploration, (b) Moment Selection, and (c) QA and Boundary Selection.}
    \label{fig:interaction_components}
\end{figure}

As depicted in \textbf{Figure}~\ref{fig:user_interface}, our system allows a seamless search experience through a three-stage process. After the submission of the query, the system retrieves the top K keyframes laid out in a grid view, with all corresponding metadata for each keyframe being included. This initial retrieval process takes advantage of the BEiT-3 model to create fine-grained visual and semantic understanding. After retrieving the first set of keyframes, the user selects a pivot keyframe to create a temporal search from these initial results. To facilitate this important step, we have optimized the process so that the user only has to enter the first portion of their query in the box. Once the user selects a pivot keyframe, they can add the remaining text of their query for the specific purpose of searching for temporal relationships. The interface also provides moment exploration, where the user can review frames that precede and follow the pivot keyframe to establish context and continuity. In the final stage of the process, as illustrated in \textbf{Figure}~\ref{fig:moment_selection},~\ref{fig:qa_boundary}, the user reviews the temporal boundaries indicated within the sequence they selected to make more pinpointed decisions about what segments to cut. Included in this step is a Question and Answering(QA) annotation process to allow users to document important observations and answer on behalf of the query's requirements.

\subsection{Data Preprocessing}
\label{subsec:data_preprocessing}
During the data preprocessing phase, we extract keyframes from the raw video data to represent its content effectively. The set of extracted keyframes is denoted as $K$, where $K_i$ refers to the keyframe located at index $i$ within the video. To achieve this, we utilize the TransNetV2\cite{soucek2024transnet} model, which is well-suited for detecting shot boundaries. For each segment with frame indices ranging from $[a,b]$, we select four keyframes \cite{le2023enhancing,phan2023doppelsearch} based on the following formula:
\begin{equation}
k_{\text{extract}} = \{ K_{a + \lfloor i \times (b-a)/3 \rfloor} \mid \forall i \in (0,1,2,3) \}
\end{equation}

\subsubsection{Keyframe Deduplication}


A common challenge in keyframe extraction is the occurrence of near-duplicate frames, where multiple consecutive keyframes contain highly similar content. To address these issues, we utilize a \textbf{near-duplicate removal} strategy  based on \textbf{perceptual hashing (pHash)} \cite{idealods2019imagededup}. The method efficiently detects and removes visually redundant keyframes by computing hash-based similarity scores. By filtering out near-duplicates within a shot detected by the previous section, we maintain a more compact yet representative set of keyframes, improving both storage efficiency and retrieval speed. 

        


To identify visual similarity between frames within arbitrary video shots, we compute the similarity between two keyframes $I_i$ and $I_j$ based on the \textit{Hamming distance} between their corresponding perceptual hash representations:

\begin{equation}
D(I_i, I_j) = \sum_{k=1}^{N} \mathbb{1} (h_i^k \neq h_j^k)
\end{equation}

Where:
\begin{itemize}
    \item $h_i$ and $h_j$ are perceptual hashes of keyframes $I_i$ and $I_j$.
    \item $N$ is the hash length (e.g., $N = 64$ for an 8×8 pHash).
    \item $\mathbb{1} (h_i^k \neq h_j^k)$ is an indicator function that counts bitwise differences.
\end{itemize}

A frame is classified as a \textbf{near-duplicate} if:

\vspace{-0.5em}
\begin{equation}
D(I_i, I_j) \leq N (1 - \tau)
\end{equation}

where $\tau$ is the similarity threshold (e.g., $\tau = 0.8$). Frames exceeding this threshold are grouped, and only one representative frame per cluster is retained.

By removing near-duplicate keyframes, we achieve less redundant data stored, leading to lower disk space requirements and faster search and retrieval.
    




\subsubsection{Feature Extractor}
In our approach, we utilize \textbf{BEiT-3} as a deep learning-based feature extractor to improve the accuracy of near-duplicate detection. BEiT-3~\cite{beit3} is a state-of-the-art vision-language model that serves as a powerful feature extractor by leveraging a transformer-based architecture to capture high-level semantic information from images. The feature extraction process involves preprocessing images into fixed-size patches, embedding them into a high-dimensional space, and passing them through multiple transformer layers to generate rich contextual representations. The final output is a compact feature vector that can be used for various tasks such as similarity comparison, image retrieval, and near-duplicate detection. By employing BeIT-3, we enhance the robustness of our feature representations, leading to more accurate and efficient visual analysis.
\subsection{Searching and Reranking}
\label{subsec:search_rerank}
\subsubsection{Searching}
In large-scale image retrieval, storing and searching through high-dimensional feature embeddings efficiently is a critical challenge. To address this, we utilize FAISS \cite{douze2024faiss}, a library designed for fast approximate nearest neighbor (ANN) search. FAISS provides scalable indexing structures that enable rapid retrieval of similar images from massive datasets. The key advantage of FAISS is its ability to handle millions to billions of vectors efficiently using optimized algorithms such as product quantization (PQ), inverted file indexes (IVF), and hierarchical navigable small world (HNSW) graphs.

In our approach, we extract global feature embeddings from images and store them in a FAISS index. During retrieval, a given query image is first converted into its feature representation, which is then used to search for the top-M nearest neighbors based on cosine similarity. This initial search serves as a candidate selection stage, providing a refined set of potential matches while maintaining computational efficiency. Using FAISS, we significantly reduce retrieval latency and memory overhead, making it feasible for large-scale datasets without sacrificing accuracy.
\subsubsection{Reranking}

A standard practice for modern moment retrieval systems is to perform coarse retrieval using global representations, and then subsequently refine the retrieval on the candidate subset using local features. However, this approach tends to be computationally intensive and it is not scalable to a large collection of videos where we extract millions of keyframes.

To achieve efficient retrieval speeds and a reduced memory footprint while preserving accuracy, we incorporate the SuperGlobal Reranking method, as introduced by Shao et al. \cite{shao2023global} in "Global Features are All You Need for Image Retrieval and Reranking". This approach uses global descriptors for both the initial and reranking steps, significantly reducing the computational overhead while maintaining accuracy. This approach leverages the Generalized Mean (GeM) pooling mechanism \cite{radenovic2018fine}, which provides the general capability for feature aggregation as described by. The global descriptor is expanded using the equation:
\begin{equation}
    f_k = \left(  \frac{1}{\mathcal{X}_k} \sum_{x \in \mathcal{X}_k} x^{p_k} \right)^{\frac{1}{p_k}}
\end{equation}
where $f_k$ is the feature to be refined, $\mathcal{X}_{k}$ represents the set of neighbors of $f_k$, and $p_k$ is a hyperparameter . We leverage two special cases of this formulation: when $p_k=1$ (average pooling) for image descriptor refinement and when $p_k \rightarrow \infty$ (max pooling) for query expansion.

SuperGlobal improves the quality of global features by refining the representation of both the query and the retrieved images. Given a query image, its global feature representation is updated in conjunction with its top-M retrieved images to generate an enhanced descriptor. This refined descriptor better captures contextual and semantic information, leading to more accurate ranking decisions.
In the reranking stage, each query image maintains both its original representation $g_q$ and an expanded representation $g_{qe}$. We then compute two sets of similarity scores:
\begin{itemize}
    \item $S_1$: Measures the similarity between the original query descriptor $g_q$ and the refined descriptors $g_{dr}$ of the database images.
    \item $S_2$: Measures the similarity between the expanded query descriptor $g_{qe}$ and the original global descriptors $g_d$.
\end{itemize}
The final reranking score $S_{final}$ is computed by averaging the two similarity scores:
\begin{equation}
    S_{final} = \frac{S_1 + S_2}{2}
\end{equation}
By incorporating SuperGlobal-based refinement, the most relevant images are ranked higher, leading to improved retrieval performance. This approach effectively enhances ranking stability while maintaining computational efficiency, making it suitable for large-scale retrieval systems.

\subsection{Temporal Search}
\label{subsec:temporal_search}

Given a user-provided pivot frame from the selection of the first process, which approximates the location of the moment described in the query, the fundamental challenge lies in determining the exact temporal extension that captures the intended action while preserving contextual coherence. Conventional similarity-based retrieval approaches often introduce difficulties in determining moment boundaries due to temporal ambiguity, where multiple visually similar frames exist in proximity to the actual moment, complicating boundary establishment. To address these issues, we propose \textbf{Adaptive Bidirectional Temporal Search}, a simple yet effective method that improves retrieval precision by jointly optimizing query relevance and temporal stability. we decompose it into two directional sub-queries targeting the start and end boundaries of the event. Our algorithm performs a backward search from the pivot to identify the start frame, and a forward search to locate the end frame. Within each direction, candidate frames are ranked using a composite score that balances how well a frame matches the query with how stable it is in its local temporal neighborhood. To quantify this stability, we introduce a confidence measure based on the standard deviation of similarity scores between each candidate frame and its nearby frames. A frame with low variance is considered temporally stable, suggesting that it belongs to a region of consistent visual content and is less likely to lie near a scene transition or visual disturbance. This property is critical for moment localization: stable frames are more likely to serve as natural semantic boundaries, acting as points of transition into or out of coherent actions or scenes. In contrast, unstable frames often occur in the middle of ongoing action, during motion blur, or at abrupt cuts—conditions under which frame-level similarity may be high but semantically misleading. By combining stability with semantic alignment, ABTS avoids selecting noisy frames and instead chooses frames that are both meaningful and consistent.

\begin{algorithm}[h]
\caption{Adaptive Bidirectional Temporal Search for Video Retrieval}
\label{alg:temporal_search}
\begin{algorithmic}[1]
\Require Query $q$, Pivot index $p$, Video dataset $\mathcal{V}$, Embedding model $\mathcal{M}$, Window sizes $\mathcal{W}$
\Ensure Start and End frame indices $(f_s, f_e)$

\State Extract $v$ from $\mathcal{V}$ corresponding to $p$
\State Compute frame embeddings $\mathcal{E}$ for $v$
\State Encode query segments: $(e_q^s, e_q^e) \gets \mathcal{M}(q)$
\State Initialize candidate sets: $\mathcal{S}, \mathcal{E} \gets \emptyset$

\For{$w \in \mathcal{W}$} 
    \State Extract local embeddings around $p$ with range $w$
    \State $s_{\text{local}}, c_s \gets \text{AdaptiveSearch}(e_q^s, \mathcal{E}_{\text{start}})$
    \State $e_{\text{local}}, c_e \gets \text{AdaptiveSearch}(e_q^e, \mathcal{E}_{\text{end}})$
    \State $\mathcal{S} \gets \mathcal{S} \cup \{(s_{\text{local}}, c_s)\}$
    \State $\mathcal{E} \gets \mathcal{E} \cup \{(e_{\text{local}}, c_e)\}$
\EndFor

\State Select $f_s = \arg\max_{(s, c) \in \mathcal{S}} c$
\State Select $f_e = \arg\max_{(e, c) \in \mathcal{E}} c$
\State Compute timestamps $(t_s, t_e)$ from $(f_s, f_e)$

\Return $(f_s, f_e, t_s, t_e)$
\end{algorithmic}
\end{algorithm}

\begin{algorithm}[h]
\caption{Adaptive Search}
\label{alg:adaptive_search}
\begin{algorithmic}
\Require Query embedding $e_q$, Frame embeddings $\mathcal{E}$, Similarity weight $\lambda_s$, Stability weight $\lambda_t$
\Ensure Best frame index $f^*$

\State Initialize similarity scores $\mathcal{C} \gets \emptyset$

\For{$i \in \{1, \dots, |\mathcal{E}|\}$} 
    \State Extract frame embedding $e_i$
    \State Identify neighboring frames $\mathcal{N}_i$
    \State Compute similarity: $s_i = Similarity(e_q, e_i)$
    \State Compute stability: $t_i = Stability(\mathcal{N}_i, e_i)$
    \State Compute confidence: $c_i = \lambda_s s_i + \lambda_t t_i$
    \State $\mathcal{C} \gets \mathcal{C} \cup \{(i, c_i)\}$
\EndFor

\State Select best frame $f^* = \arg\max_{(i, c) \in \mathcal{C}} c$
\Return $f^*$
\end{algorithmic}
\end{algorithm}

The \textbf{Algorithm} ~\ref{alg:temporal_search} describes the implementation of the adaptive bidirectional temporal search, designed to locate the most relevant temporal segment given a query. The algorithm operates by first identifying the video that contains the pivot index $p$ in the video corpus $\mathcal{V}$ and computing frame-wise embedding $\mathcal{E}$ using a pre-trained embedding model $\mathcal{M}$. The query $q$ is then split into two sub-queries, each describing the anticipated start and end of the target moment, and subsequently encoded by $\mathcal{M}$ to obtain the corresponding embeddings $e^{s}_{q}$ and $e^{e}_{q}$. Since a moment in this workshop ranges from 2–20 seconds, the search is performed over multiple temporal window sizes $\mathcal{W}$, extracting local embeddings around the pivot frame $p$. As the pivot can algorithmically represent either boundary, we set the window list to be 10 seconds, 15 seconds, and 20 seconds. The adaptive search algorithm (\textbf{Algorithm}~\ref{alg:adaptive_search}) is then applied separately to locate the optimal start and end frames based on similarity and stability scores. The highest confidence frame indices are selected as the final segment boundaries ($f_s$, $f_e$), which are subsequently assigned to timestamps ($t_s$, $t_e$), providing precise temporal localization of the retrieved moment.

The key component of the moment localization process lies in \textbf{Algorithm}~\ref{alg:adaptive_search}, which selects the most relevant frames by computing a confidence-weighted score that integrates both semantic similarity and temporal stability. Given a query embedding $e_q$ and a set of candidate frame embeddings $\mathcal{E}$, the algorithm iterates through each frame and determines the score based on two complementary measures. The similarity score $s_i$ quantifies how closely a frame embedding aligns with the query and is computed as:
\begin{equation}
    s_i = \frac{e_q \cdot e_i}{||e_q|| ||e_i||}
\end{equation}
While similarity alone captures semantic alignment, it is often insufficient due to visual noise or abrupt scene transitions. To migrate this, the algorithm incorporates a stability score $t_i$, which determines how consistent a frame is within its local temporal neighborhood $\mathcal{N}_i$. This is formulated as:
\begin{equation}
t_i = 1 - min(1,2~\cdot~ \sigma(\{e_j ~ \cdot ~ e_i|j \in \mathcal{N}_i\}) )    
\end{equation}
Where $\sigma(\cdot)$ denotes the standard deviation of cosine similarities among neighboring frames. This formulation ensures that frames belonging to stable temporal regions receive higher scores, reducing the likelihood of selecting outliers, blurred, or transition frames [Repeat Temporal Stability]. The final confidence score, $c_i$, is obtained  by computing the weighted similarity and stability score with hyperparameters $\lambda_s$ and $\lambda_t$:
\begin{equation}
    c_i = \lambda_s s_i + \lambda_t t_i
\end{equation}
By jointly optimizing for semantic relevance and temporal coherence, the adaptive search mechanism robustly selects the most reliable frame, ensuring precise localization of the retrieved video moment.

\section{Experimental results}
In this section, we evaluate our interactive video retrieval system on two tasks: \textbf{Known-Item Search (KIS)} and \textbf{Video Question Answering (QA)}. Each case study highlights the impact of key system components such as reranking and temporal search, providing qualitative insights into system performance, strengths, and limitations. We showcase a representative example for each task that best demonstrates the system’s effectiveness in real-world scenarios.

\subsection{Know-Item Search task}
\subsubsection{With Reranking}
 Our system first retrieves the most relevant keyframes, followed by the proposed reranking strategy to prioritize semantically relevant results.

Example Query: \textit{`Begin with the gleaming trophy on display at the center of the perfectly manicured field, then transition to the broadcast team and commentators preparing for the live coverage. Show the packed stadium with thousands of enthusiastic fans in team colors, followed by dramatic moments when pyrotechnics light up the boundary during a celebration. End with intimate team moments as players huddle together in their distinctive uniforms, showing both the New Zealand team in blue and the Australian team in green and gold.`}

\begin{figure}[t]
    \centering
    \begin{subfigure}{0.45\textwidth}
        \centering
        \fbox{\includegraphics[width=\linewidth]{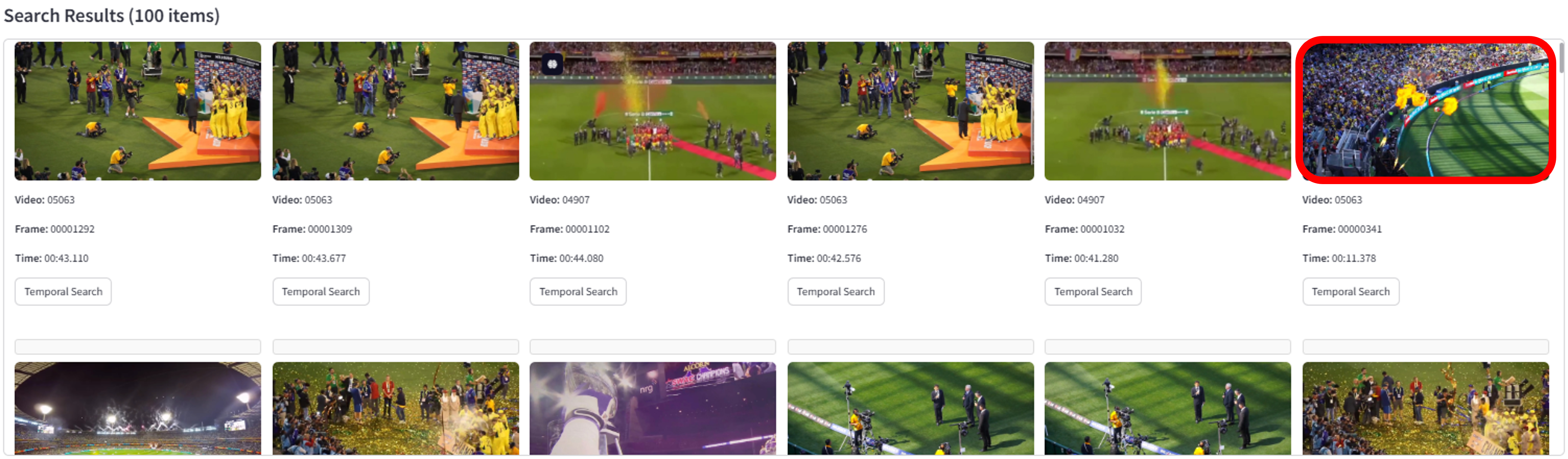}}
        \caption{Before using reranking}
    \end{subfigure}
    \hfill
    \begin{subfigure}{0.45\textwidth}
        \centering
        \fbox{\includegraphics[width=\linewidth]{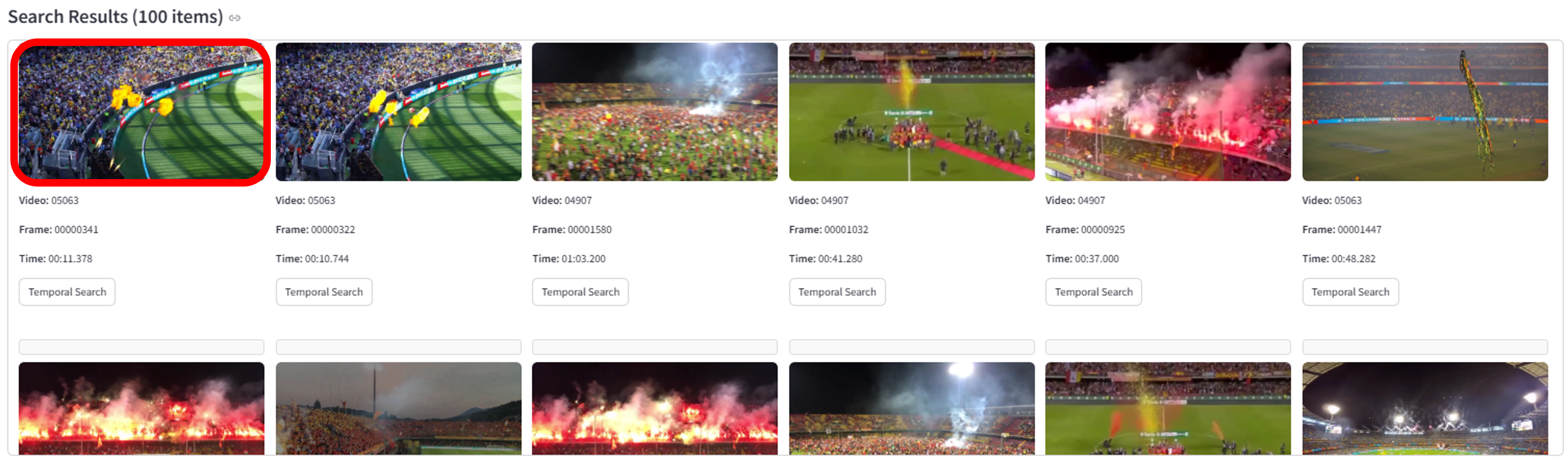}}
        \caption{After using reranking}
    \end{subfigure}
    \caption{Demonstration of our reranking function's effectiveness in retrieving frames most matching to the query.}
    \label{fig:rerank_images}
\end{figure}

As shown in \textbf{Figure}~\ref{fig:rerank_images}, without reranking, the target frame ranked in the top 6. With our proposed reranking, it is promoted to the top 3, demonstrating improved relevance and ranking accuracy. 

The effect of reranking is especially evident when applied to complex, multi-step queries such as the one describing a full broadcast sequence—from trophy presentation to live coverage to stadium-wide celebration. As evidenced by the visual results, the initial retrieval surfaces visually similar frames (e.g., stadium scenes, large crowds), yet semantically inconsistent with the requested celebration phase. Several top-ranked results depict unrelated moments, such as pre-game camera setups or post-match crowd dispersals. After reranking, the system correctly promotes the key frame featuring pyrotechnics lighting up the boundary, precisely matching the “dramatic moments” described in the query. Additionally, the surrounding top frames shift to show synchronized crowd celebrations, reinforcing alignment with the event’s peak moment. This demonstrates the system’s ability to distinguish between narrative phases and to prioritize results that match scene-specific actions and context—a crucial capability when retrieving fine-grained moments within a larger event timeline.

\subsubsection{With Temporal Search}
Beyond frame retrieval, users should be able to localize the full video segment corresponding to the query. Our adaptive temporal search identifies the start and end boundaries by incorporating both semantic similarity and temporal stability, while also capturing key spatial and temporal aspects of the video content. Spatial scenarios are demonstrated in the following query: \textit{`Capture a journey through a valley with shots from a moving vehicle, showing the turquoise river winding between steep cliffs and terraced fields. Frame the dramatic mountain range in the background with billowing white clouds embracing their peaks while keeping the lush green vegetation in the foreground. Include perspectives from bridge crossings that frame the river below with metal railings in the foreground. End with a lingering shot from the center of the bridge.`}

\begin{figure}[t]
    \centering
    \begin{subfigure}{0.45\textwidth}
        \centering
        \fbox{\includegraphics[width=\linewidth]{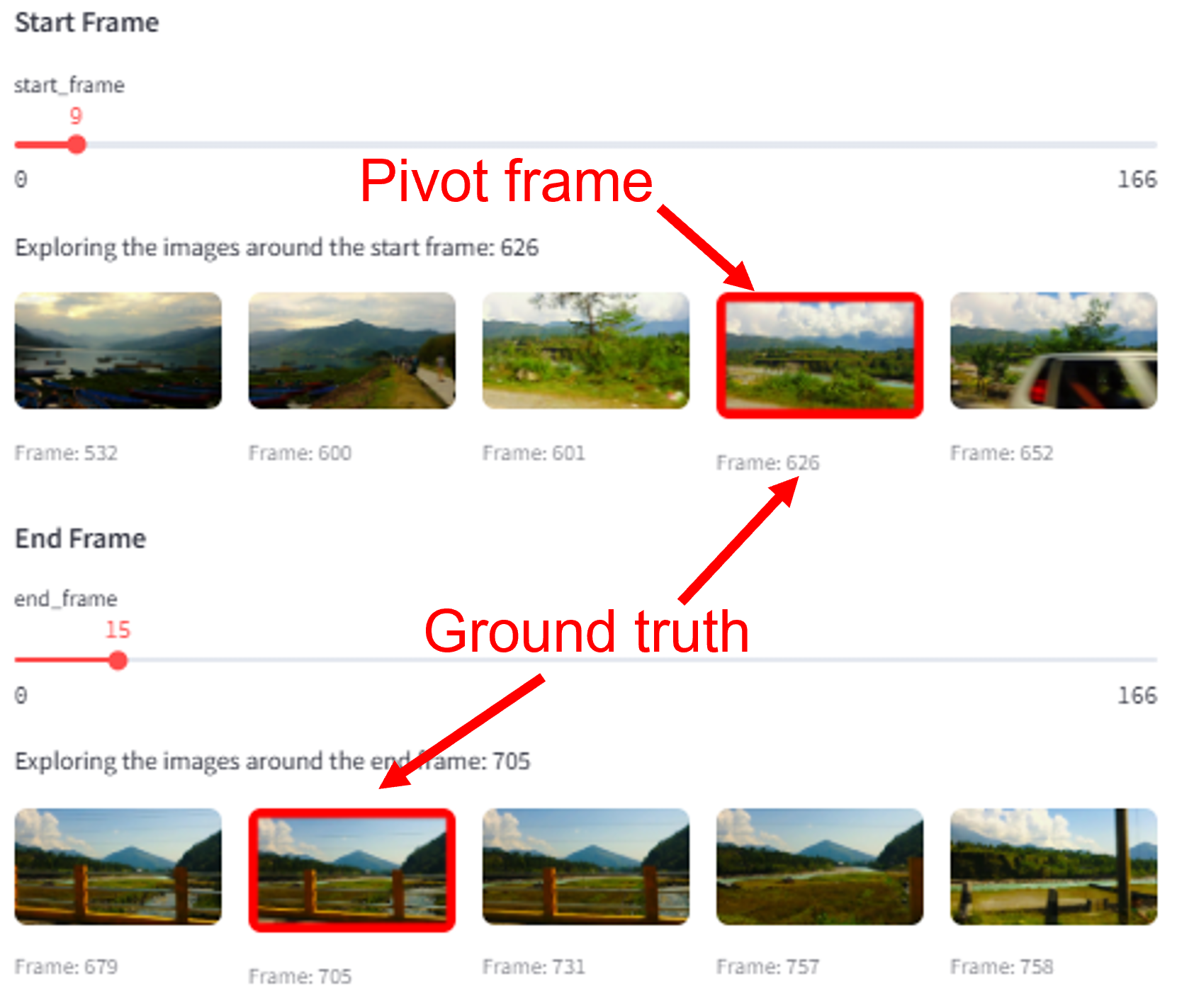}}
    \end{subfigure}
    \caption{Demonstration of our temporal search function's ability to recognize and interpret complex spatial compositions across video sequences. The selected frames reflect accurate alignment with the query’s described layout, capturing foreground, background, and motion cues to support precise moment localization.}
    \label{fig:spatial_temporal_images}
\end{figure}

This query helps identify complex spatial relationships in the scene: from foreground elements (river, mountain), and background components (clouds) to motion indicators (vehicle movement). As in \textbf{Figure}~\ref{fig:spatial_temporal_images}, our system demonstrates strong spatial understanding by selecting frames that accurately reflect the query’s described scene composition. The chosen start frame (Frame 626) captures a clear view of the turquoise river, fence, and distant mountains, aligning well with the spatial layout outlined in the query, with the surrounding frames offering consistent visual context. Notably, the end frame (Frame 705) precisely matches the description of the last sentence of the query. A stable shot of the bridge is selected over local keyframes, demonstrating the capability of the system to detect boundaries that best align with both the semantic and visual content of the query. 

Beyond retrieval accuracy, our system offers an interactive interface that enhances user experience. As shown in \textbf{Figure}~\ref{fig:spatial_temporal_images}, the interface presents surrounding frames alongside the selected start and end frames, allowing users to see and verify the moment segment. By enabling users to validate scene continuity and spatial consistency in real-time, the system facilitates more precise and user-aligned moment localization.

Temporal factors are also demonstrated in the following query: \textit{`The images depict a high-stakes poker game in dramatic close-up shots. The camera focuses on hands managing poker chips and cards on a red felt table. Hands adorned with gold rings adjust stacks of colorful chips across the table. The camera focuses intensely on fingers gripping the edge of face-down cards. With deliberate slowness, the hand turns over its hidden treasure. In the final, climactic shot, two aces are revealed—a pair of powerful pocket rockets that signal a game-changing moment about to unfold.`}

\begin{figure}[t]
    \centering
    \begin{subfigure}{0.45\textwidth}
        \centering
        \fbox{\includegraphics[width=\linewidth]{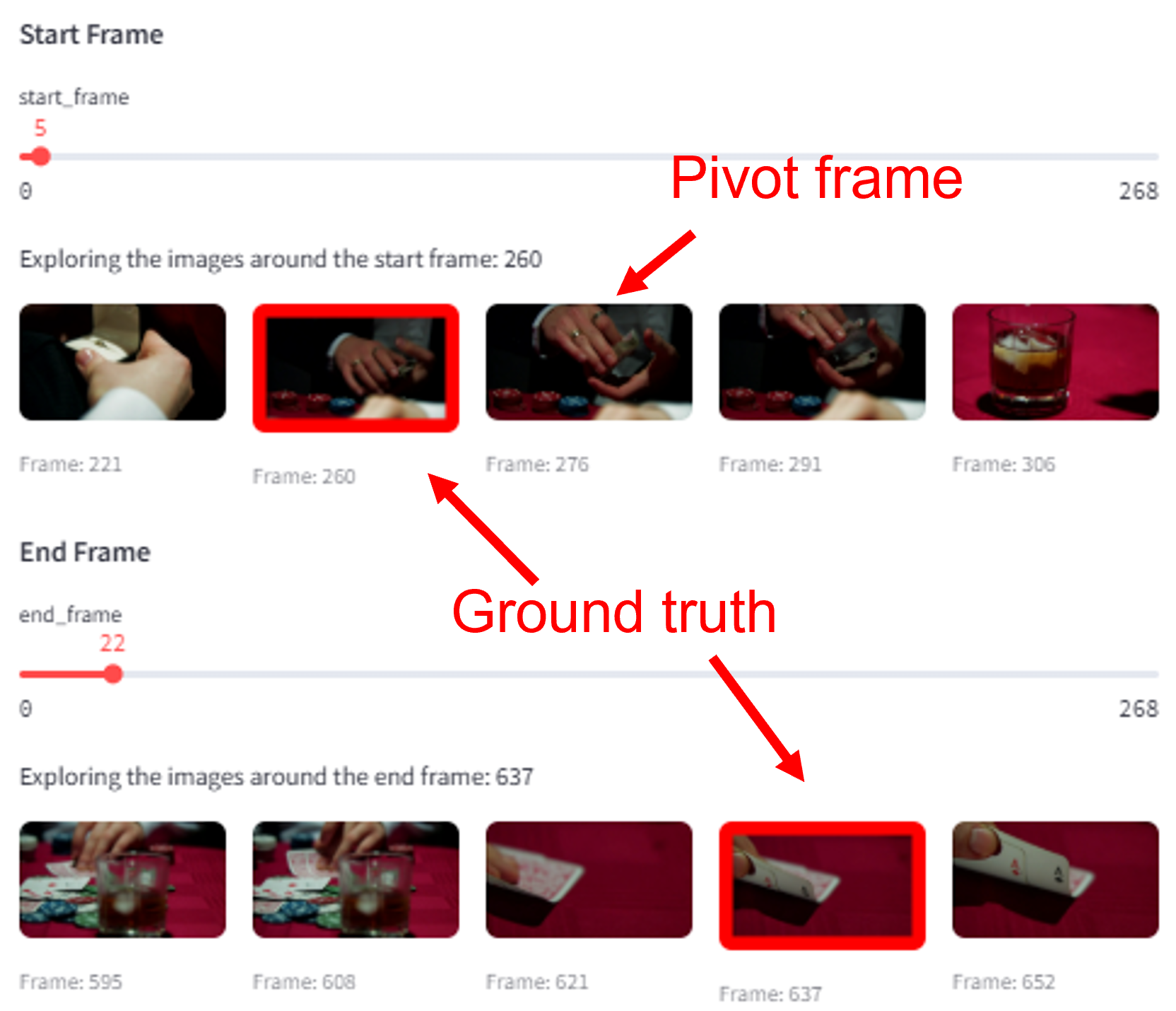}}
    \end{subfigure}
    \caption{In this demonstration, we highlight our search function's advanced temporal recognition capabilities, specifically in the context of a high-stakes poker game scenario. }
    \label{fig:temporal_images}
\end{figure}

As in \textbf{Figure}~\ref{fig:temporal_images}, our system effectively captures temporal narratives by identifying the precise start and end boundaries aligned with evolving events. The selected segment follows a clear trend: from preparatory actions like handling poker chips and drinks (Frame 260) to the climactic reveal of two aces (Frame 637), matching the described tension and resolution in the query. The surrounding frames provide smooth transitions that reinforce the story progression, demonstrating the capability of the system to track not only visual content but also the underlying temporal dynamics. These results demonstrate the effectiveness of our approach in handling complex temporal narratives, rather than static visual descriptions alone.

In addition to accurate boundary localization, our system demonstrates robust temporal reasoning by identifying frames that align with both semantic transitions and temporal coherence. As shown in \textbf{Figure}~\ref{fig:temporal_images}, while multiple frames share similar visual elements (e.g., red felt, poker chips, hands), the system effectively selects Frame 637 as the endpoint, precisely capturing the climactic card reveal. This choice reflects more than just visual similarity—it highlights the ability of the model to detect semantic shifts in the narrative, from tension-building actions to the moment of resolution. 

\subsection{Question Answering task}
For the \textbf{Question Answering} task, we select the following query: \textit{`A man wearing dark trousers and jacket, a scarf and a horse mask walks down a path. There is a wooden bench on the right, cars and houses in the background, and litter on the path and grass. He shakes his head and puts his hand to the head. The camera follows him as he walks to another wooden bench and sits down. Which writer is quoted at the start of this video`}?


\begin{figure}[t]
    \centering
    \begin{subfigure}{0.45\textwidth}
        \centering
       \fbox{\includegraphics[width=\linewidth]{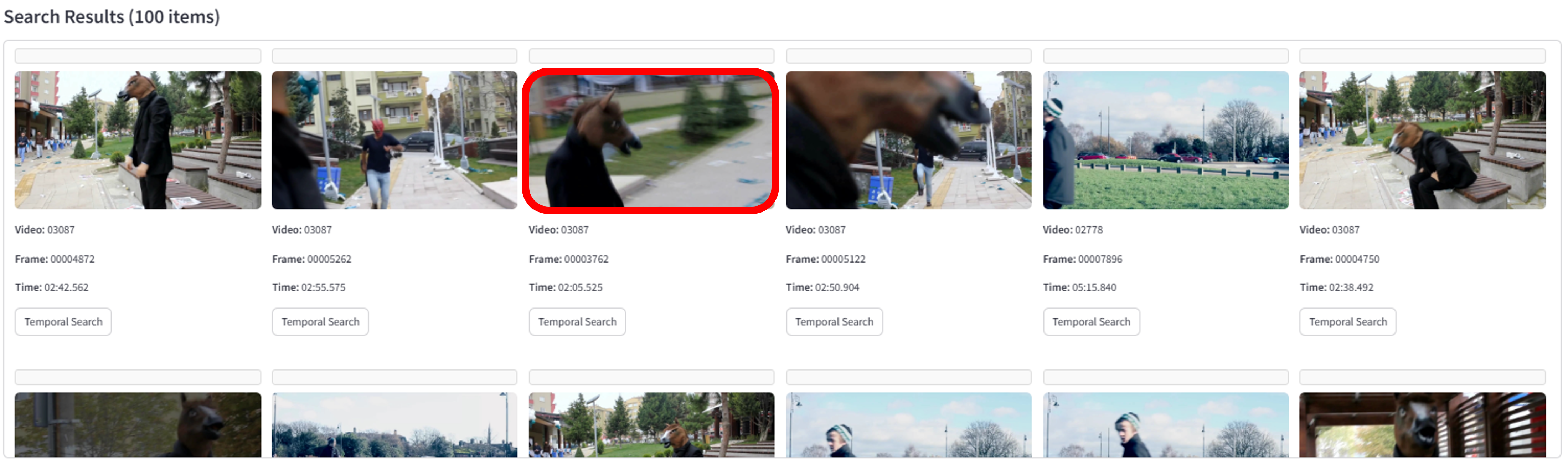}}
    \end{subfigure}
    \caption{Demonstration of accurate retrieval for a specific video segment matching a detailed natural language query. This example highlights the system’s ability to precisely localize complex, multi-step actions and visual context, showcasing its effectiveness in handling fine-grained moment retrieval within a large video corpus.}
    \label{fig:question_answering}
\end{figure}


As in \textbf{Figure}~\ref{fig:question_answering}, the system retrieves the appropriate segment based on the event sequence. Users can refine the result by selecting anchor frames and adjusting temporal boundaries. The final answer—displayed in a subsequent keyframe—is the quoted text:
\textit{“Vicdanımız yanılmaz bir yargıçtır, biz onu öldürmedikçe - Balzac”}.
\label{sec:experiment}
\section{Conclusion}
\label{sec:conclusion}

In conclusion, our comprehensive experimental evaluation demonstrates the robust capabilities of the Interactive VCMR framework across two critical tasks: Known-Item Search (KIS) and Video Question Answering(QA) By demonstrating promising performance across diverse scenarios, our approach proves the ability to capture not only the visual similarity but semantic progression and temporal stability yield a significant performance boost in video moment retrieval system. These findings reinforce the potential of intelligent, context-aware retrieval systems to transform how we interact with and navigate large video repositories.

\newpage
{
    \small
    \bibliographystyle{ieeenat_fullname}
    \bibliography{main}

\begin{thebibliography}{81}
\providecommand{\natexlab}[1]{#1}
\providecommand{\url}[1]{\texttt{#1}}
\expandafter\ifx\csname urlstyle\endcsname\relax
  \providecommand{\doi}[1]{doi: #1}\else
  \providecommand{\doi}{doi: \begingroup \urlstyle{rm}\Url}\fi

\bibitem[aiv()]{aivietnamVit}
{A}{I} {VIETNAM} --- aivietnam.edu.vn.

\bibitem[Agarwal et~al.(2020)Agarwal, Balasubramanian, Choudhury, et~al.]{agarwal2020active}
A. Agarwal, V. Balasubramanian, S. Choudhury, et~al.
\newblock Active learning for interactive video retrieval.
\newblock In \emph{Proceedings of the NeurIPS Workshop on Human-in-the-Loop Learning}, 2020.

\bibitem[Akbari et~al.(2021)Akbari, Yuan, Qian, Chuang, Fu, Cui, et~al.]{akbari2021vatt}
H. Akbari, L. Yuan, R. Qian, W.C. Chuang, Y. Fu, Y. Cui, et~al.
\newblock Vatt: Transformers for multimodal self-supervised learning from raw video, audio and text.
\newblock In \emph{Advances in Neural Information Processing Systems (NeurIPS)}, 2021.

\bibitem[Alam et~al.(2022)Alam, Semedo, Ahmad, et~al.]{alam2022revisiting}
F. Alam, J. Semedo, W.U. Ahmad, et~al.
\newblock Revisiting video retrieval with textual feedback.
\newblock In \emph{Proceedings of the Conference on Empirical Methods in Natural Language Processing (EMNLP)}, pages 7788--7800, 2022.

\bibitem[Amato et~al.(2021)Amato, Bolettieri, Falchi, Gennaro, Messina, Vadicamo, and Vairo]{Amato2021VISIONE}
Giuseppe Amato, Paolo Bolettieri, Fabrizio Falchi, Claudio Gennaro, Nicola Messina, Lucia Vadicamo, and Claudio Vairo.
\newblock {VISIONE at Video Browser Showdown 2021}.
\newblock In \emph{Proc. of the 27th Int. Conf. on MultiMedia Modeling (MMM)}, pages 473--478. Springer, Cham, 2021.

\bibitem[Arandjelovi{\'c} and Zisserman(2012{\natexlab{a}})]{arandjelovic2012query}
R. Arandjelovi{\'c} and A. Zisserman.
\newblock Query expansion for image retrieval.
\newblock In \emph{Proceedings of the European Conference on Computer Vision (ECCV)}, pages 168--181, 2012{\natexlab{a}}.

\bibitem[Arandjelovi{\'c} and Zisserman(2012{\natexlab{b}})]{arandjelovic2012three}
Relja Arandjelovi{\'c} and Andrew Zisserman.
\newblock Three things everyone should know to improve object retrieval.
\newblock In \emph{2012 IEEE conference on computer vision and pattern recognition}, pages 2911--2918. IEEE, 2012{\natexlab{b}}.

\bibitem[Bain et~al.(2021)Bain, Nagrani, Tzinis, and Zisserman]{bain2021frozen}
M. Bain, A. Nagrani, E. Tzinis, and A. Zisserman.
\newblock Frozen in time: A joint video and image encoder for end-to-end retrieval.
\newblock In \emph{Proceedings of the IEEE/CVF International Conference on Computer Vision (ICCV)}, pages 1728--1738, 2021.

\bibitem[Bao et~al.(2023)Bao, Kong, Shao, Ng, Er, and Kot]{anonymous2023vidmorp}
Peijun Bao, Chenqi Kong, Zihao Shao, Boon~Poh Ng, Meng~Hwa Er, and Alex~C. Kot.
\newblock Vid-morp: Video moment retrieval pretraining from unlabeled videos in the wild.
\newblock \emph{arXiv preprint arXiv:2412.00811}, 2023.

\bibitem[Chen et~al.(2023)]{chen2023semantic}
S. Chen et~al.
\newblock Semantic reconstruction network for video moment retrieval.
\newblock In \emph{Proceedings of ACM International Conference on Multimedia}, 2023.

\bibitem[Chen et~al.(2022)Chen, Huang, Zhang, Wang, Dong, and Yu]{chen2022few}
Z. Chen, W. Huang, L. Zhang, J. Wang, J. Dong, and Z. Yu.
\newblock Few-shot video retrieval via interaction-aware memory alignment.
\newblock In \emph{Proceedings of the European Conference on Computer Vision (ECCV)}, pages 624--642, 2022.

\bibitem[Christel and Hauptmann(2005)]{christel2005improving}
M. Christel and A.G. Hauptmann.
\newblock Improving user interaction in video retrieval.
\newblock In \emph{Proceedings of the International Conference on Image and Video Retrieval}, pages 61--70, 2005.

\bibitem[Cui et~al.(2022)]{cui2022video}
R. Cui et~al.
\newblock Video moment retrieval from text queries via single frame annotation.
\newblock \emph{arXiv preprint arXiv:2204.09409}, 2022.

\bibitem[Das et~al.(2017)Das, Kottur, Gupta, et~al.]{das2017learning}
A. Das, S. Kottur, K. Gupta, et~al.
\newblock Learning cooperative visual dialog agents with deep reinforcement learning.
\newblock In \emph{Proceedings of the IEEE International Conference on Computer Vision (ICCV)}, pages 2951--2960, 2017.

\bibitem[Douze et~al.(2024)Douze, Guzhva, Deng, Johnson, Szilvasy, Mazar{\'e}, Lomeli, Hosseini, and J{\'e}gou]{douze2024faiss}
Matthijs Douze, Alexandr Guzhva, Chengqi Deng, Jeff Johnson, Gergely Szilvasy, Pierre-Emmanuel Mazar{\'e}, Maria Lomeli, Lucas Hosseini, and Herv{\'e} J{\'e}gou.
\newblock The faiss library.
\newblock \emph{arXiv preprint arXiv:2401.08281}, 2024.

\bibitem[Fang et~al.(2023)Fang, Cheng, Gan, et~al.]{fang2023clip2video}
H. Fang, Y. Cheng, Z. Gan, et~al.
\newblock Clip2video: Mastering video-text retrieval via image clip.
\newblock In \emph{Proceedings of the Advances in Neural Information Processing Systems (NeurIPS)}, 2023.

\bibitem[Gao and Xu(2021)]{gao2021fast}
J. Gao and C. Xu.
\newblock Fast video moment retrieval.
\newblock In \emph{Proceedings of the International Conference on Computer Vision (ICCV)}, 2021.

\bibitem[Gao et~al.(2018)Gao, Ge, Chen, and Nevatia]{gao2018motion}
J. Gao, R. Ge, K. Chen, and R. Nevatia.
\newblock Motion-appearance co-memory networks for video question answering.
\newblock In \emph{Proceedings of the IEEE Conference on Computer Vision and Pattern Recognition (CVPR)}, pages 6576--6585, 2018.

\bibitem[Hou et~al.(2024)Hou, Pang, Shen, and Cheng]{hou2024event}
Danyang Hou, Liang Pang, Huawei Shen, and Xueqi Cheng.
\newblock Event-aware video corpus moment retrieval.
\newblock \emph{arXiv preprint arXiv:2402.13566}, 2024.

\bibitem[Jain et~al.(2016)Jain, Snoek, and Worring]{jain2016query}
M. Jain, C.G.M. Snoek, and M. Worring.
\newblock Query-adaptive video summarization via quality-aware relevance estimation.
\newblock \emph{IEEE Transactions on Multimedia}, 19\penalty0 (12):\penalty0 2665--2676, 2016.

\bibitem[Jain et~al.(2019)Jain, Lennan, John, and Tran]{idealods2019imagededup}
Tanuj Jain, Christopher Lennan, Zubin John, and Dat Tran.
\newblock Imagededup.
\newblock \url{https://github.com/idealo/imagededup}, 2019.

\bibitem[Jang et~al.(2017)Jang, Song, Kim, Yu, Kim, and Kim]{jang2017tgifqa}
Y. Jang, Y. Song, C.D. Kim, Y. Yu, Y. Kim, and G. Kim.
\newblock Tgif-qa: Toward spatio-temporal reasoning in visual question answering.
\newblock In \emph{Proceedings of the IEEE Conference on Computer Vision and Pattern Recognition (CVPR)}, pages 2758--2766, 2017.

\bibitem[Jang et~al.(2019)Jang, Song, Kim, Yu, Kim, and Kim]{jang2019video}
Y. Jang, Y. Song, C.D. Kim, Y. Yu, Y. Kim, and G. Kim.
\newblock Video question answering with spatio-temporal reasoning.
\newblock \emph{International Journal of Computer Vision}, 127\penalty0 (10):\penalty0 1385--1412, 2019.

\bibitem[Jiang et~al.(2020)Jiang, Chen, Lin, Zhao, and Gao]{jiang2020divide}
J. Jiang, Z. Chen, H. Lin, X. Zhao, and Y. Gao.
\newblock Divide and conquer: Question-guided spatio-temporal contextual attention for video question answering.
\newblock In \emph{Proceedings of the AAAI Conference on Artificial Intelligence}, pages 11101--11108, 2020.

\bibitem[Jones et~al.(2024)]{jones2024lddetr}
T. Jones et~al.
\newblock Ld-detr: Loop decoder detection transformer for video moment retrieval and highlight detection.
\newblock In \emph{Proceedings of the IEEE/CVF Conference on Computer Vision and Pattern Recognition (CVPR)}, 2024.

\bibitem[J{\'o}nsson et~al.(2020)J{\'o}nsson, Khan, Koelma, Rudinac, Worring, and Zah{\'a}lka]{Jonsson2020Exquisitor}
Bj{\"o}rn~{\TH}{\'o}r J{\'o}nsson, Omar~Shahbaz Khan, Dennis~C. Koelma, Stevan Rudinac, Marcel Worring, and Jan Zah{\'a}lka.
\newblock {Exquisitor at the Video Browser Showdown 2020}.
\newblock In \emph{Proc. of the 26th Int. Conf. on MultiMedia Modeling (MMM)}, pages 796--802. Springer, Cham, 2020.

\bibitem[Jung et~al.(2022)Jung, Choi, Kim, Kim, and Zhang]{jung2022modal}
M. Jung, S. Choi, J. Kim, J.H. Kim, and B.T. Zhang.
\newblock Modal-specific pseudo query generation for video corpus moment retrieval.
\newblock In \emph{Proceedings of the 2022 Conference on Empirical Methods in Natural Language Processing}, pages 7769--7781, Abu Dhabi, UAE, 2022.

\bibitem[Law et~al.(2018)Law, Guan, Uijlings, and Hospedales]{law2018vision}
M.T.H. Law, Y. Guan, J. Uijlings, and T.M. Hospedales.
\newblock Vision through conversations: Real-time object grounding using guided dialogue.
\newblock In \emph{Proceedings of the IEEE Conference on Computer Vision and Pattern Recognition (CVPR)}, pages 7350--7359, 2018.

\bibitem[Le-Quynh et~al.(2023)Le-Quynh, Nguyen, Quang-Hoang, Dinh, Nguyen, Ngo, and An]{le2023enhancing}
Minh-Dung Le-Quynh, Anh-Tuan Nguyen, Anh-Tuan Quang-Hoang, Van-Huy Dinh, Tien-Huy Nguyen, Hoang-Bach Ngo, and Minh-Hung An.
\newblock Enhancing video retrieval with robust clip-based multimodal system.
\newblock In \emph{Proceedings of the 12th International Symposium on Information and Communication Technology}, pages 972--979, 2023.

\bibitem[Lee et~al.(2021)Lee, Kim, Shin, and Yoo]{lee2021video}
J.Y. Lee, J.W. Kim, J. Shin, and C. Yoo.
\newblock Video-grounded dialog via transformer with multi-level semantic alignment.
\newblock In \emph{Proceedings of the AAAI Conference on Artificial Intelligence}, pages 196--204, 2021.

\bibitem[Lei et~al.(2021)Lei, Li, Zhou, Gan, Berg, Bansal, et~al.]{lei2021clipbert}
J. Lei, L. Li, L. Zhou, Z. Gan, T.L. Berg, M. Bansal, et~al.
\newblock Less is more: Clipbert for video-and-language learning via sparse sampling.
\newblock In \emph{Proceedings of the IEEE/CVF Conference on Computer Vision and Pattern Recognition (CVPR)}, pages 7331--7341, 2021.

\bibitem[Leibetseder et~al.(2020)Leibetseder, M{\"u}nzer, Primus, Kletz, and Sch{\"o}ffmann]{Leibetseder2020DiveXplore}
Andreas Leibetseder, Bernd M{\"u}nzer, J{\"u}rgen Primus, Sabrina Kletz, and Klaus Sch{\"o}ffmann.
\newblock {diveXplore 4.0: The ITEC Deep Interactive Video Exploration System at VBS2020}.
\newblock In \emph{Proc. of the 26th Int. Conf. on MultiMedia Modeling (MMM)}, pages 753--759. Springer, Cham, 2020.

\bibitem[Li et~al.(2022)Li, Wei, Tian, Xu, Wen, and Hu]{li2022learning}
G. Li, Y. Wei, Y. Tian, C. Xu, J.R. Wen, and D. Hu.
\newblock Learning to answer questions in dynamic audio-visual scenarios.
\newblock In \emph{Proceedings of the IEEE/CVF Conference on Computer Vision and Pattern Recognition (CVPR)}, pages 19108--19118, 2022.

\bibitem[Li et~al.(2020)Li, Chen, Cheng, Gan, Yu, and Liu]{li2020hero}
L. Li, Y. Chen, Y. Cheng, Z. Gan, L. Yu, and J. Liu.
\newblock Hero: Hierarchical encoder for video–language omni-representation pre-training.
\newblock In \emph{Proceedings of the Empirical Methods in Natural Language Processing (EMNLP)}, pages 2046--2065, 2020.

\bibitem[Li et~al.(2019)Li, Song, Gao, Liu, et~al.]{li2019beyond}
X. Li, J. Song, L. Gao, X. Liu, et~al.
\newblock Beyond rnns: Positional self-attention with co-attention for video question answering.
\newblock In \emph{Proceedings of the AAAI Conference on Artificial Intelligence}, pages 8658--8665, 2019.

\bibitem[Lin et~al.(2022)]{lin2022weakly}
D. Lin et~al.
\newblock Weakly-supervised video moment retrieval via semantic completion network.
\newblock In \emph{European Conference on Computer Vision (ECCV)}, 2022.

\bibitem[Liu et~al.(2021)Liu, Xu, Qin, Zhang, Wang, and Shao]{liu2021dynamic}
J. Liu, Y. Xu, J. Qin, Y. Zhang, X. Wang, and L. Shao.
\newblock Dynamic concept interaction for visual dialog.
\newblock In \emph{Proceedings of the IEEE/CVF Conference on Computer Vision and Pattern Recognition (CVPR)}, pages 3629--3638, 2021.

\bibitem[Liu et~al.(2023{\natexlab{a}})Liu, Lin, Lin, et~al.]{liu2023hyper}
M. Liu, Z. Lin, J. Lin, et~al.
\newblock Hyper-personalized video search with few-shot user feedback.
\newblock In \emph{Proceedings of the 16th ACM International Conference on Web Search and Data Mining (WSDM)}, pages 1075--1084, 2023{\natexlab{a}}.

\bibitem[Liu et~al.(2023{\natexlab{b}})]{liu2023boundary}
M. Liu et~al.
\newblock Boundary-aware bilinear network for dense video moment retrieval.
\newblock \emph{IEEE Transactions on Image Processing}, 2023{\natexlab{b}}.

\bibitem[Loko{\v{c}} et~al.(2019)Loko{\v{c}}, Koval{\v{c}}{\'i}k, Sou{\v{c}}ek, Moravec, Bodn{\'a}r, and {\v{C}}ech]{Lokoc2019VIRET}
Jakub Loko{\v{c}}, Gregor Koval{\v{c}}{\'i}k, Tom{\'a}{\v{s}} Sou{\v{c}}ek, Jaroslav Moravec, Jan Bodn{\'a}r, and P{\v{r}}emysl {\v{C}}ech.
\newblock {VIRET Tool Meets NasNet}.
\newblock In \emph{Proc. of the 25th Int. Conf. on MultiMedia Modeling (MMM)}, pages 597--601. Springer, Cham, 2019.

\bibitem[Lokoč et~al.(2019)Lokoč, Kovalčík, Münzer, Schoeffmann, Bailer, Gasser, et~al.]{lokoc2019interactive}
J. Lokoč, G. Kovalčík, B. Münzer, K. Schoeffmann, W. Bailer, R. Gasser, et~al.
\newblock Interactive search or sequential browsing? a detailed analysis of the video browser showdown 2018.
\newblock \emph{ACM Transactions on Multimedia Computing, Communications, and Applications (TOMM)}, 15\penalty0 (1):\penalty0 Article 29, 2019.

\bibitem[Lokoč et~al.(2023)Lokoč, Andreadis, Bailer, Duane, Gurrin, et~al.]{lokoc2023interactive}
J. Lokoč, S. Andreadis, W. Bailer, A. Duane, C. Gurrin, et~al.
\newblock Interactive video retrieval in the age of effective joint embedding deep models: Lessons from the 11th vbs.
\newblock \emph{Multimedia Systems}, 29:\penalty0 3481--3504, 2023.

\bibitem[Mao et~al.(2022)Mao, Jiang, Wang, Feng, Lyu, Liu, et~al.]{mao2022dynamic}
J. Mao, W. Jiang, X. Wang, Z. Feng, Y. Lyu, H. Liu, et~al.
\newblock Dynamic multistep reasoning based on video scene graph for video question answering.
\newblock In \emph{Proceedings of the NAACL-HLT}, pages 3894--3904, 2022.

\bibitem[Miech et~al.(2020)Miech, Alayrac, Laptev, Sivic, and Zisserman]{miech2020end}
A. Miech, J.B. Alayrac, I. Laptev, J. Sivic, and A. Zisserman.
\newblock End-to-end learning of visual representations from uncurated instructional videos.
\newblock In \emph{Proceedings of the IEEE/CVF Conference on Computer Vision and Pattern Recognition (CVPR)}, pages 9879--9888, 2020.

\bibitem[Mun et~al.(2022)]{mun2022local}
S. Mun et~al.
\newblock Local-global video-text interaction algorithm for video moment retrieval.
\newblock \emph{IEEE Transactions on Pattern Analysis and Machine Intelligence}, 2022.

\bibitem[Ngo et~al.(2024)Ngo, Lam, Nguyen, Dinh, and Choi]{ngo2024dual}
Ba~Hung Ngo, Ba~Thinh Lam, Thanh~Huy Nguyen, Quang~Vinh Dinh, and Tae~Jong Choi.
\newblock Dual dynamic consistency regularization for semi-supervised domain adaptation.
\newblock \emph{IEEE Access}, 2024.

\bibitem[Nguyen et~al.(2025{\natexlab{a}})Nguyen, Lam, Truong, Duong, and Dinh]{nguyen5109180mv}
Huy~Thanh Nguyen, Thinh~Ba Lam, Toan Thai~Ngoc Truong, Thang~Dinh Duong, and Vinh~Quang Dinh.
\newblock Mv-trams: An efficient tumor region-adapted mammography synthesis under multi-view diagnosis.
\newblock \emph{Available at SSRN 5109180}, 2025{\natexlab{a}}.

\bibitem[Nguyen et~al.(2025{\natexlab{b}})Nguyen, Vu, Duong, Duong, Nguyen, and Dinh]{nguyen2025enhancing}
Khoi~Anh Nguyen, Linh~Yen Vu, Thang~Dinh Duong, Thuan~Nguyen Duong, Huy~Thanh Nguyen, and Vinh~Quang Dinh.
\newblock Enhancing vietnamese vqa through curriculum learning on raw and augmented text representations.
\newblock \emph{arXiv preprint arXiv:2503.03285}, 2025{\natexlab{b}}.

\bibitem[Nguyen et~al.(2024{\natexlab{a}})Nguyen, Tran, and Quang-Hoang]{nguyen2024improvinggeneralizationvisualreasoning}
Tien-Huy Nguyen, Quang-Khai Tran, and Anh-Tuan Quang-Hoang.
\newblock Improving generalization in visual reasoning via self-ensemble, 2024{\natexlab{a}}.

\bibitem[Nguyen et~al.(2024{\natexlab{b}})Nguyen, Tran, Tran, Phan-Nguyen, and Nguyen]{10661057}
Tho-Quang Nguyen, Huu-Loc Tran, Tuan-Khoa Tran, Huu-Phong Phan-Nguyen, and Tien-Huy Nguyen.
\newblock Fa-yolov9: Improved yolov9 based on feature attention block.
\newblock In \emph{2024 International Conference on Multimedia Analysis and Pattern Recognition (MAPR)}, pages 1--6, 2024{\natexlab{b}}.

\bibitem[Nguyen et~al.(2024{\natexlab{c}})Nguyen, Nguyen, Nguyen, Do, and Dinh]{nguyen2024emotic}
Xuan-Bach Nguyen, Hoang-Thien Nguyen, Thanh-Huy Nguyen, Nhu-Tai Do, and Quang~Vinh Dinh.
\newblock Emotic masked autoencoder on dual-views with attention fusion for facial expression recognition.
\newblock In \emph{Proceedings of the IEEE/CVF Conference on Computer Vision and Pattern Recognition}, pages 4784--4792, 2024{\natexlab{c}}.

\bibitem[Papadopoulos et~al.(2017)Papadopoulos, Clarke, Keller, et~al.]{papadopoulos2017click}
D. Papadopoulos, A. Clarke, F. Keller, et~al.
\newblock Training object class detectors with click supervision.
\newblock In \emph{Proceedings of the IEEE Conference on Computer Vision and Pattern Recognition (CVPR)}, pages 5794--5802, 2017.

\bibitem[Park et~al.(2021)Park, Lee, and Sohn]{park2021bridge}
J. Park, J. Lee, and K. Sohn.
\newblock Bridge to answer: Structure-aware graph interaction network for video question answering.
\newblock In \emph{Proceedings of the IEEE/CVF Conference on Computer Vision and Pattern Recognition (CVPR)}, pages 15526--15535, 2021.

\bibitem[Phan Nguyen~Huu et~al.(2023)Phan Nguyen~Huu, Tran~Dinh, Tran Kim~Ngoc, Tran~Duc, Le~Hoang, Nguyen~Huu, Phan~The, and Pham]{phan2023doppelsearch}
Phong Phan Nguyen~Huu, Khoa Tran~Dinh, Ngan Tran Kim~Ngoc, Luong Tran~Duc, Phuc Le~Hoang, Quyen Nguyen~Huu, Duy Phan~The, and Van-Hau Pham.
\newblock Doppelsearch: A novel approach to content-based video retrieval for ai challenge hcmc 2023.
\newblock In \emph{Proceedings of the 12th International Symposium on Information and Communication Technology}, pages 916--922, 2023.

\bibitem[Philbin et~al.(2007)Philbin, Chum, Isard, Sivic, and Zisserman]{philbin2007object}
James Philbin, Ondrej Chum, Michael Isard, Josef Sivic, and Andrew Zisserman.
\newblock Object retrieval with large vocabularies and fast spatial matching.
\newblock In \emph{2007 IEEE conference on computer vision and pattern recognition}, pages 1--8. IEEE, 2007.

\bibitem[Radenovi{\'c} et~al.(2018)Radenovi{\'c}, Tolias, and Chum]{radenovic2018fine}
Filip Radenovi{\'c}, Giorgos Tolias, and Ond{\v{r}}ej Chum.
\newblock Fine-tuning cnn image retrieval with no human annotation.
\newblock \emph{IEEE transactions on pattern analysis and machine intelligence}, 41\penalty0 (7):\penalty0 1655--1668, 2018.

\bibitem[Rui et~al.(1998)Rui, Huang, Ortega, and Mehrotra]{rui1998relevance}
Y. Rui, T.S. Huang, M. Ortega, and S. Mehrotra.
\newblock Relevance feedback: a power tool for interactive content-based image retrieval.
\newblock \emph{IEEE Transactions on Circuits and Systems for Video Technology}, 8\penalty0 (5):\penalty0 644--655, 1998.

\bibitem[Sauter et~al.(2020)Sauter, Parian, Gasser, Heller, Rossetto, and Schuldt]{Sauter2020Vitrivr}
Loris Sauter, Mahnaz~Amiri Parian, Ralph Gasser, Silvan Heller, Luca Rossetto, and Heiko Schuldt.
\newblock {Combining Boolean and Multimedia Retrieval in vitrivr for Large-Scale Video Search}.
\newblock In \emph{Proc. of the 26th Int. Conf. on MultiMedia Modeling (MMM)}, pages 760--765. Springer, Cham, 2020.

\bibitem[Seo et~al.(2021)Seo, Lee, Oh, Kim, and Kim]{seo2021reinforced}
Y. Seo, K. Lee, T. Oh, I. Kim, and B. Kim.
\newblock Reinforced interactive video retrieval with semantic memory.
\newblock In \emph{Proceedings of the 29th ACM International Conference on Multimedia (ACM MM)}, pages 234--242, 2021.

\bibitem[Shao et~al.(2023)Shao, Chen, Karpur, Cui, Araujo, and Cao]{shao2023global}
Shihao Shao, Kaifeng Chen, Arjun Karpur, Qinghua Cui, Andr{\'e} Araujo, and Bingyi Cao.
\newblock Global features are all you need for image retrieval and reranking.
\newblock In \emph{Proceedings of the IEEE/CVF International Conference on Computer Vision}, pages 11036--11046, 2023.

\bibitem[Smeaton et~al.(2006)Smeaton, Over, and Kraaij]{smeaton2006evaluation}
A.F. Smeaton, P. Over, and W. Kraaij.
\newblock Evaluation campaigns and trecvid.
\newblock In \emph{Proceedings of the ACM International Workshop on Multimedia Information Retrieval}, pages 321--330, 2006.

\bibitem[Song et~al.(2018)Song, Shi, Chen, and Han]{song2018multi}
X. Song, Y. Shi, X. Chen, and Y. Han.
\newblock Explore multi-step reasoning in video question answering.
\newblock In \emph{Proceedings of the 26th ACM International Conference on Multimedia}, pages 239--247, 2018.

\bibitem[Soucek and Lokoc(2024)]{soucek2024transnet}
Tom{\'a}s Soucek and Jakub Lokoc.
\newblock Transnet v2: An effective deep network architecture for fast shot transition detection.
\newblock In \emph{Proceedings of the 32nd ACM International Conference on Multimedia}, pages 11218--11221, 2024.

\bibitem[Tan et~al.(2021)Tan, Yuan, and Ordonez]{tan2021instance}
Fuwen Tan, Jiangbo Yuan, and Vicente Ordonez.
\newblock Instance-level image retrieval using reranking transformers.
\newblock In \emph{proceedings of the IEEE/CVF international conference on computer vision}, pages 12105--12115, 2021.

\bibitem[Wang et~al.(2023{\natexlab{a}})]{wang2023tree}
J. Wang et~al.
\newblock Tree-structured lstm for video moment retrieval.
\newblock In \emph{Proceedings of the IEEE/CVF Winter Conference on Applications of Computer Vision}, 2023{\natexlab{a}}.

\bibitem[Wang et~al.(2023{\natexlab{b}})]{wang2023graph}
M. Wang et~al.
\newblock Graph neural networks for video moment retrieval with cross-modal relations.
\newblock \emph{IEEE Transactions on Multimedia}, 2023{\natexlab{b}}.

\bibitem[Wang et~al.(2023{\natexlab{c}})Wang, Bao, Dong, Bjorck, Peng, Liu, Aggarwal, Mohammed, Singhal, Som, and Wei]{beit3}
Wenhui Wang, Hangbo Bao, Li Dong, Johan Bjorck, Zhiliang Peng, Qiang Liu, Kriti Aggarwal, Owais~Khan Mohammed, Saksham Singhal, Subhojit Som, and Furu Wei.
\newblock Image as a foreign language: {BEiT} pretraining for vision and vision-language tasks.
\newblock In \emph{Proceedings of the IEEE/CVF Conference on Computer Vision and Pattern Recognition}, 2023{\natexlab{c}}.

\bibitem[Wu et~al.(2022)]{wu2022reinforcement}
J. Wu et~al.
\newblock Reinforcement learning for temporal grounding of natural language in untrimmed videos.
\newblock In \emph{Proceedings of AAAI Conference on Artificial Intelligence}, 2022.

\bibitem[Wu et~al.(2023)Wu, Cao, Bai, Zeng, Chen, Nie, and Zhang]{wu2023empirical}
Mengxia Wu, Min Cao, Yang Bai, Ziyin Zeng, Chen Chen, Liqiang Nie, and Min Zhang.
\newblock An empirical study of frame selection for text-to-video retrieval.
\newblock \emph{arXiv preprint arXiv:2311.00298}, 2023.

\bibitem[Wu et~al.(2021)Wu, Garcia, Otani, Chu, Nakashima, and Takemura]{wu2021transferring}
T. Wu, N. Garcia, M. Otani, C. Chu, Y. Nakashima, and H. Takemura.
\newblock Transferring domain-agnostic knowledge in video question answering.
\newblock \emph{arXiv preprint arXiv:2110.13395}, 2021.

\bibitem[Xiao et~al.(2022)Xiao, Zhou, Chua, and Yan]{xiao2022video}
J. Xiao, P. Zhou, T.S. Chua, and S. Yan.
\newblock Video graph transformer for video question answering.
\newblock In \emph{Proceedings of the European Conference on Computer Vision (ECCV)}, 2022.

\bibitem[Xu et~al.(2017)Xu, Zhao, Xiao, Wu, Zhang, He, et~al.]{xu2017gradually}
D. Xu, Z. Zhao, J. Xiao, F. Wu, H. Zhang, X. He, et~al.
\newblock Video question answering via gradually refined attention over appearance and motion.
\newblock In \emph{Proceedings of the 25th ACM International Conference on Multimedia (ACM MM)}, pages 1645--1653, 2017.

\bibitem[Yang et~al.(2021)Yang, Miech, Sivic, Laptev, and Schmid]{yang2021justask}
A. Yang, A. Miech, J. Sivic, I. Laptev, and C. Schmid.
\newblock Just ask: Learning to answer questions from millions of narrated videos.
\newblock In \emph{Proceedings of the IEEE/CVF International Conference on Computer Vision (ICCV)}, pages 1686--1697, 2021.

\bibitem[Yang et~al.(2022)Yang, Miech, Sivic, Laptev, and Schmid]{yang2022zeroshot}
A. Yang, A. Miech, J. Sivic, I. Laptev, and C. Schmid.
\newblock Zero-shot video question answering via frozen bidirectional language models.
\newblock \emph{Advances in Neural Information Processing Systems (NeurIPS)}, 2022.

\bibitem[Yi et~al.(2020)Yi, Gan, Li, Kohli, et~al.]{yi2020clevrer}
K. Yi, C. Gan, Y. Li, P. Kohli, et~al.
\newblock Clevrer: Collision events for video representation and reasoning.
\newblock In \emph{Proceedings of the International Conference on Learning Representations (ICLR)}, 2020.
\newblock Paper no. 4527.

\bibitem[Yin et~al.(2024)Yin, Ma, Cao, Wang, Chen, and Jiang]{yin2024text}
Zhihui Yin, Ye Ma, Xipeng Cao, Bo Wang, Quan Chen, and Peng Jiang.
\newblock Text-video multi-grained integration for video moment montage.
\newblock \emph{arXiv preprint arXiv:2412.09276}, 2024.

\bibitem[Zellers et~al.(2021)Zellers, Lu, Hessel, Yu, Park, Cao, et~al.]{zellers2021merlot}
R. Zellers, X. Lu, J. Hessel, Y. Yu, J.S. Park, J. Cao, et~al.
\newblock Merlot: Multimodal neural script knowledge models.
\newblock \emph{Advances in Neural Information Processing Systems (NeurIPS)}, 34, 2021.

\bibitem[Zhang et~al.(2021)Zhang, Sun, Jing, Nan, Zhen, Zhou, and Goh]{zhang2021video}
Hao Zhang, Aixin Sun, Wei Jing, Guoshun Nan, Liangli Zhen, Joey~Tianyi Zhou, and Rick Siow~Mong Goh.
\newblock Video corpus moment retrieval with contrastive learning.
\newblock In \emph{Proceedings of the 44th International ACM SIGIR Conference on Research and Development in Information Retrieval}, pages 685--695, 2021.

\bibitem[Zhang et~al.(2025)Zhang, Song, Duan, Wang, Chang, and Yang]{zhang2025video}
Long Zhang, Peipei Song, Zhangling Duan, Shuo Wang, Xiaojun Chang, and Xun Yang.
\newblock Video corpus moment retrieval with query-specific context learning and progressive localization.
\newblock \emph{IEEE Transactions on Circuits and Systems for Video Technology}, 2025.

\bibitem[Zhang et~al.(2024)]{zhang2024momentgpt}
X. Zhang et~al.
\newblock Moment-gpt: Zero-shot video moment retrieval via off-the-shelf multimodal large language models.
\newblock In \emph{Proceedings of the IEEE/CVF Conference on Computer Vision and Pattern Recognition}, 2024.

\bibitem[Zhao et~al.(2023)]{zhao2023timeloc}
L. Zhao et~al.
\newblock Timeloc: A unified end-to-end framework for precise timestamp localization in long videos.
\newblock \emph{arXiv preprint arXiv:2311.09509}, 2023.

\end{thebibliography}

}


\end{document}